%% file: rl4rl_ws.tex
\documentclass{article}

\usepackage{microtype}
\usepackage{graphicx}
\usepackage{booktabs} % for professional tables

\usepackage{hyperref}

\newcommand{\targetform}{\theta_f}
\newcommand{\targetroughness}{\theta_\sigma}

\usepackage{caption}
\usepackage{subcaption}

\usepackage[accepted]{icml2021}

\icmltitlerunning{Monte Carlo Tree Search for high precision manufacturing}
\usepackage{siunitx}
\usepackage{amsmath}
\usepackage{amsfonts}
\usepackage{amssymb}
\usepackage{lscape}

\usepackage[utf8]{inputenc}
\usepackage{pgfplots}
\DeclareUnicodeCharacter{2212}{−}
\usepgfplotslibrary{groupplots,dateplot}
\usetikzlibrary{patterns,shapes.arrows}
\pgfplotsset{compat=newest}
\begin{document}

\twocolumn[
\icmltitle{%Monte Carlo Tree Search for high precision manufacturing
Optimization of high precision manufacturing by Monte Carlo Tree Search
}



\icmlsetsymbol{equal}{*}

\begin{icmlauthorlist}
\icmlauthor{Dorina Weichert*}{iais1,iais2}
\icmlauthor{Felix Horchler*}{bonn}
\icmlauthor{Alexander Kister}{iais1}
\icmlauthor{Marcus Trost}{iof}
\icmlauthor{Johannes Hartung}{iof}
\icmlauthor{Stefan Risse}{iof}
\end{icmlauthorlist}

\icmlaffiliation{iais2}{Fraunhofer Center for Machine Learning}
\icmlaffiliation{iais1}{Fraunhofer Institute for Intelligent Analysis and Information Systems IAIS, Schloss Birlinghoven, 53757 Sankt Augustin, Germany}
\icmlaffiliation{bonn}{Institute of Computer Science, Bonn University, 53115 Bonn, Germany}
\icmlaffiliation{iof}{Fraunhofer Institue for Applied Optics and Precision Engineering IOF, 07745 Jena, Germany}

\icmlcorrespondingauthor{Dorina Weichert}{dorina.weichert@iais.fraunhofer.de}

\icmlkeywords{Monte Carlo Tree Search, POMDP, high precision manufacturing}

\vskip 0.3in
]



\printAffiliationsAndNotice{\icmlEqualContribution} % otherwise use the standard text.

\begin{abstract}
Monte Carlo Tree Search (MCTS) has shown its strength for a lot of deterministic and stochastic examples, but literature lacks reports of applications to real world industrial processes.
Common reasons for this are that there is no efficient simulator of the process available or there exist problems in applying MCTS to the complex rules of the process.
In this paper, we apply MCTS for optimizing a high-precision manufacturing process that has stochastic and partially observable outcomes. We make use of an expert-knowledge-based simulator and adapt the MCTS default policy to deal with the manufacturing process.
\end{abstract}

\section{Introduction}
Reinforcement learning has demonstrated its strengths in real-life applications such as games \cite{silver2017mastering, Mnih2013PlayingAW}, robot control \cite{Kober2013robotics} and optimization of energy consumption \cite{data_center}. We present its use in high precision manufacturing, where we have to deal with two challenges: a) we have to introduce high level expert knowledge into the optimization approach and b) we have to deal with a partially observable production process that has stochastic outcomes.
%{\color{green} Maybe there is a way around the "standard" initial phrase: Possibly by saying RL is improved by adressing practical challanges like lake of human examples silver, ...  }

The processing results of real-life manufacturing processes always contain an element of randomness, e.g. the manufacturing tolerances.
In order to reach a given product quality, determined by a product tolerance, (e.g. the allowed deviation from a design), engineers have to choose manufacturing processes with an appropriate manufacturing tolerance.
\begin{figure*}
     \centering
     \begin{subfigure}[b]{0.3\textwidth}
         \centering
         \includegraphics[width=\textwidth]{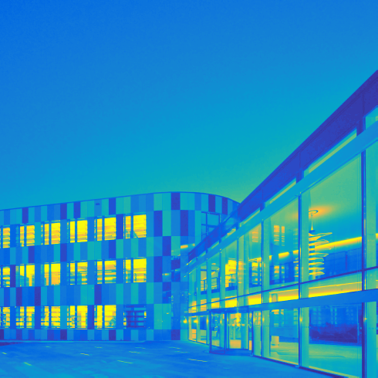}
         \caption{ideal image without degration}
     \end{subfigure}
     \hfill
     \begin{subfigure}[b]{0.3\textwidth}
         \centering
         \includegraphics[width=\textwidth]{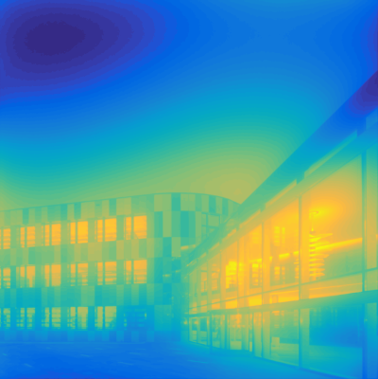}
         \caption{impact of scattering due to roughness}
     \end{subfigure}
     \hfill
     \begin{subfigure}[b]{0.3\textwidth}
         \centering
         \includegraphics[width=\textwidth]{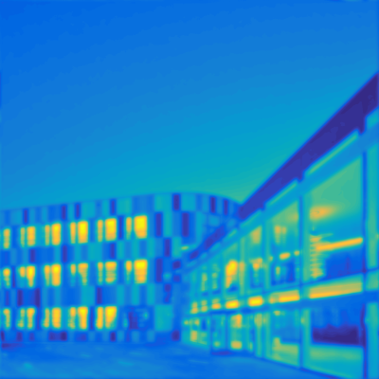}
         \caption{aberration due to shape deviation}
     \end{subfigure}
        \caption{Comparison of the impact of different surface imperfections on the imaging quality.}
        \label{fig:manufacturing_errors}
\end{figure*}

In common manufacturing, the product tolerances are much greater than the manufacturing tolerances. They can be treated as deterministic, as the noise of the manufacturing outcomes influence the processing result only to a minor extent. In this paper, we deal with the less common case of high precision manufacturing: here, the manufacturing tolerances of the different processing steps are in the range of the product tolerance.
As the manufacturing outcomes vary, the chain of manufacturing steps has to be adapted. Hence, the state-of-the-art is to perform a processing step, determine its outcome by taking a measurement and then to choose the next manufacturing step based on this current measurement value and the history of processing steps.
As a consequence, we distinguish two types of manufacturing steps: \textit{processing steps} change the manufacturing outcome, while \textit{measurement steps} return information about the manufactured component.

The chain of process steps, and the associated production time, highly depends on the expert knowledge of the responsible process engineer. To become more independent of this individual factor, it is very attractive to find a time-optimal chain of process steps. In addition, measurement steps are costly regarding the overall manufacturing time and do not change the manufacturing outcome. Thus, reducing the number of measurement steps is a main optimization goal.

Monte Carlo Tree Search (MCTS) (see \cite{Browne2012survey} for a detailed review) is a reinforcement learning method that is well suited for problems in the high precision manufacturing domain, as it can solve optimization problems with large state-spaces. In this paper, we use an extension of the method for partially observable processes, Partially Observable Monte-Carlo Planning (POMCP) \cite{Silver2010POMDP}, that is suitable to treat the noisy manufacturing outcomes (both measurement and processing steps return noisy results) and the fact that the product quality can only be observed in measurement steps. 

To optimize the production time, we consider the following three points. First, we use expert knowledge to model the manufacturing and measurement in form of a a particle-based simulator and we calibrate this model to experimental data. Second, we adapt POMCP to learn fast manufacturing stategies. Our adaptions correspond to incorporating process specific requirements (e.g. the measurements take time, but enlarge the knowledge about the product state) and expert knowledge into the default policy. The adapted POMCP expresses the optimal manufacturing strategies as a "tree of actions", branching at observations (an example is shown on the right hand side of figure~\ref{fig:transform_tree}). Third, to validate our results, we consider a real live application scenario, where we compare our tree of actions against the process chain used by process engineers. Our results were generated for the production of a special product, namely a highly polished mirror, but are transferable to other scheduling problems where manufacturing tolerances are in the range of the product tolerances.

Overall, our work documents a full approach from the implementation of an expert-knowledge based simulator with stochastic outcomes to optimization via POMCP with a knowlegde-based adapted default policy.

\section{Related Work}
Reinforcement learning is successfully used in the area of control problems, for example, robotics \cite{Kober2013robotics}, energy management \cite{data_center}, and optics \cite{wankerl2020parameterized}. 
In domains that can be represented as trees of sequential decisions, 
MCTS is commonly used~\cite{Browne2012survey}.
It has become particularly well known for its application in the game of computer Go \cite{silver2017mastering} and many other examples such as chess, shogi and several Atari games \cite{Browne2012survey,Schrittwieser2020MasteringAG}. 
But its general concept makes it interesting for practical scheduling and optimization problems \cite{MCTS_scheduling,Browne2012survey}. In this paper, we deal with a simplified scheduling problem where only one component is processed and the correct order of manufacturing steps has to be found. But the results of the manufacturing steps are noisy and only partially observable. Partial observability is a known challenge in real-world applications of Reinforcement Learning \cite{Dulac2020}.

The main problem for time optimized high precision manufacturing is to balance the advantage of measurements steps, namely the additional information obtained by taking this measurement, with their disadvantages, namely the time added to the total manufacturing time due to this measurement.
We do this by adapting the POMCP algorithm by \cite{Silver2010POMDP}, that solves partially observable Markov decision processes (POMDP) by an MCTS-based approach. POMCP is an online approach, which means that it needs access to a real system. Our approach is an offline version, which does not need access to a real system.
In practice, this is a strength of our approach, as our optimization result is a "tree of actions", branching at observations. This means, that we return recommendations for actions depending on the real-life observation made.

\section{Description of the manufacturing process}
\label{sec:component}
To manufacture a highly polished mirror, a substrate is processed by different manufacturing steps %to affect 
each affecting two quality metrics of the mirror: its shape \(f\) \cite{ISOshape} and its roughness \(\sigma\) \cite{ISOroughness}, each being parts of a frequency range of deviations to the optical surface description from the design. The low-spatial frequency errors contribute directly into the shape deviation, while the mid-spatial and high-spatial frequency errors contribute to the roughness \cite{Pertermann:18}. The deviations are typically non-zero, since every manufacturing process introduces errors. 

In figure~\ref{fig:manufacturing_errors}, the impact of both quantities is illustrated for an imaging optic. Roughness deviations cause scattering leading to a loss of contrast, while the resolution is still high. Shape deviations on the other hand cause aberration, leading to a lower resolution. 

To reach the manufacturing goal, consisting of a target roughness and shape accuracy, a process engineer can apply two types of manufacturing processes: processing and measurement steps. 

Processing steps change the roughness or the shape of the mirror or both. In the use-case, we assume that magneto-rheological finishing (MRF) impacts both the shape and roughness \cite{beier2013fabrication}. For computer controlled polishing (CCP) the impact depends on the machining parameters. Therefore, we introduce three types of CCP: CCP1, CCP2 and CCP3, each referring to a defined set of machining parameters. CCP1 is used for large form deviations and only impacts the shape, while CCP2 and CCP3 impact both roughness and shape and are used when the quality is already high.  
Except for CCP3 all of the process steps can be applied multiple times, i.e. making multiple cycles with the manufacturing tool. The first cycle is the most effective. When the number of cycles increases, the improvement of each cycle declines. 

Measurement steps are used to determine the actual values of the mirror metrics. While the roughness is measured by a scattered light sensor (SLS) \cite{Herffurth2019AssessingSI,Trost2014InSA}; the shape is examined by either deflectometry, interferometry or profilometry, each having a different measurement accuracy and duration.

All the individual manufacturing steps have a general setup time, including tempering and cleaning, and a specific processing time, which depends on the size \(A\) of the mirror.

\section{Simulation of the manufacturing process}
\label{sec:simulator}
In the simulation of the manufacturing process, several requirements must be met. 
First, the simulation model has to represent the noisiness and the partial observability of the manufacturing process. In addition, the parameters of the simulation model have to be chosen such that they are in line with expert knowledge and empirical data.

\subsection{Structure of the simulation}
\begin{table*}[t]
  \caption{Overview of the processing and measurement steps. If not stated otherwise, values indicate the changes of the objectives, not the objective values.}
  \label{tab:manu_steps}
  \centering
  \begin{tabular}{llll}
    \toprule
    Name     & new shape \(f_{t+1}\)   & new roughness $\sigma_{t+1}$ & Duration / min \\
    \midrule
    \multicolumn{4}{l}{\textit{Processing steps}}                   \\
    \cmidrule(r){1-4}
    \textbf{CCP1} & & &  \\
    cycle 1 & $ f_t \cdot \Delta f; \Delta f \sim \mathcal{N}\left(0.9, 0.01 \alpha\right)$  &$\sigma_t$& $120 + 540/7850 \cdot A$  \\
    cycle 2-5 & $ f_t \cdot \Delta f; \Delta f \sim \mathcal{N}\left(0.925, 0.01\alpha\right)$  &$\sigma_t$& $ 540/7850 \cdot A$  \\
    cycle \(>\) 5 & $ f_t \cdot \Delta f; \Delta f \sim \mathcal{N}\left(0.95, 0.01\alpha\right)$  &$\sigma_t$& $ 540/7850 \cdot A$ \\
    \midrule
    \textbf{CCP2} &&&\\
    $\sigma_t < \SI{0.6}{nm}$ &$f_t$ &$\sigma_{t+1} \sim \mathcal{N}\left(0.6, 0.03\alpha\right)$& \\
    cycle 1 & $ f_t \cdot \Delta f; \Delta f \sim \mathcal{N}\left(1.03, 0.1\alpha\right)$ & $ \sigma_t \cdot \Delta \sigma; \Delta \sigma \sim \mathcal{N}\left(0.61, 0.03\alpha\right)$ & $ 120 + 480/7850 \cdot A$   \\
    cycle \(>\) 1 & $f_t \cdot \Delta f; \Delta f \sim \mathcal{N}\left(1.03, 0.1\alpha\right)$ & $ \sigma_t \cdot \Delta \sigma; \Delta \sigma \sim \mathcal{N}\left(0.88, 0.03\alpha\right)$ & $ 480/7850 \cdot A$   \\
    \midrule
    \textbf{CCP3} &&&\\
    $\sigma_t < \SI{0.1}{nm}$ &$f_t$&$\sigma_{t+1} \sim \mathcal{N}\left(0.1, 0.005\alpha\right)$& \\
    cycle 1 & $ f_t \cdot \Delta f; \Delta f \sim \mathcal{N}\left(1.05, 0.05\alpha\right)$  & $\sigma_t \cdot \Delta \sigma; \Delta \sigma \sim \mathcal{N}(0.75, 0.03\alpha)$ & $120 + 480/7850 \cdot A$  \\
    \midrule
    \textbf{MRF} &&&\\
    $\sigma_t < \SI{2}{nm}$ &$f_t$&$\sigma_{t+1} \sim \mathcal{N}\left(2, 0.01\alpha\right)$&\\
    cycle 1 & $ f_t \cdot \Delta f; \Delta f \sim \mathcal{N}\left(0.6, 0.01 \alpha \right)$  & $\sigma_t \cdot (0.995 - g\left(\alpha, \Delta f, f_t, \sigma_t \right))$ & $120 + 480/{7850} \cdot A$  \\
    cycle 2 & $ f_t \cdot \Delta f; \Delta f \sim \mathcal{N}\left(0.74, 0.01 \alpha \right)$  & $\sigma_t \cdot (0.995 - g\left(\alpha, \Delta f, f_t, \sigma_t \right))$ & $ 240/{7850} \cdot A$  \\
    cycle \(>\) 2 & $ f_t \cdot \Delta f; \Delta f \sim \mathcal{N}\left(0.81, 0.01\alpha \right)$  & $\sigma_t \cdot (0.995 - g\left(\alpha, \Delta f, f_t, \sigma_t \right))$ & $ 300/7850 \cdot A$  \\
    \midrule
    \multicolumn{4}{l}{\textit{Measurement steps}}                   \\
    \cmidrule(r){1-4}
    \textbf{SLS} & & $\sigma_{t+1} \sim \mathcal{N}(\sigma_t, 0.05 \alpha)$ & $h_1(A)$ \\
    \midrule
    \textbf{Deflectometry} & $f_{t+1} \sim \mathcal{N}(f_t, 0.1\alpha)$ &  & $120 + h_2(\sqrt{A}) + A/200 $ \\
    \midrule
    \textbf{Interferometry} & $f_{t+1} \sim \mathcal{N}(f_t, 0.15\alpha)$ &  & $120 + h_3(\sqrt{A}) + 150 + 7.5 $ \\
    \midrule
    \textbf{Profilometry} & $f_{t+1} \sim \mathcal{N}(f_t, 0.2\alpha)$ &  & $120 + h_4(\sqrt{A}) + 0.024\cdot A $ \\
    \bottomrule
  \end{tabular}
\end{table*}
We treat the manufacturing as a POMDP. In a Markov Decision Process (MDP), an agent moves between states \(s, s' \in \mathcal{S}\) by taking actions 
\(a \in \mathcal{A}\) with transition probabilities 
\(\mathcal{P}_{ss'}^a = Pr\left(s_{t+1} = s' \vert s_t = s, a_t = a\right)\) and receives a reward 
\(r\) from the reward function 
\(\mathcal{R}_s^a= \mathbb{E}\left[r_{t + 1} \vert s_t = s, a_t = a\right]\). 
While the agent in an MDP, can observe the state directly, this is not possible for the agent in a POMDP. In a POMDP, the agent receives a noisy observation 
\(o \in \mathcal{O}\) with observation probability 
\(Z_{s'o}^a = Pr\left(o_{t+1} = o \vert s_{t+1} = s', a_t = a\right)\).
We assume the agent knows the transition and observation probabilities. So given the historic sequence of actions and observations 
\(h_t = \left\{a_1, o_1, ..., a_t, o_t\right\}\) and an initial belief for the state $s_{0}$ in form of a distribution \(Pr\left( s_{0} = s\right)\), the agent is able to determine at every point in time $t$ its belief about the current state: we formally define this belief as \(B\left(s, h\right) = Pr(s_t = s \vert h_t = h)\).
After a sequence of actions of length $T$ is executed, the agent's actions are evaluated by the final return 
\(R_t = \sum_{l = t}^{T} r_l\), that is the sum of immediate rewards $r_l$.

In our case, the state \(s\) corresponds to a tuple of the actual quality metrics and, for each type of processing step, a counter for the number of cycles that have been executed. 
We use the counter to represent the decrease in manufacturing effectiveness over the number of cycles.
Without the counter, the decrease in manufacturing effectiveness over the number of cycles would result in the process not satisfying the Markov property.

The manufacturing steps are treated as actions \(a\). We assume two types of actions: measurement steps leading to observations \(o\) and  processing steps that change the quality metrics \((f, \sigma)\). Both kind of steps raise the counter for the action that has been taken. The measurement steps return observations \(o\) with normally distributed observation probabilities. Performing a measurement is modeled by sampling from a normal distribution with state dependent mean and standard deviation, see table \ref{tab:manu_steps}. 
%{\color{cyan}

For the processing steps, we implement the transition probabilities as followed.
We encoded reset bounds for the application of some process steps, e.g. if after application of MRF the roughness is below \SI{2}{nm}, the roughness value is reset to a value sampled from \(\mathcal{N}(2, 0.01 \alpha)\), see table \ref{tab:manu_steps}. 
For all manufacturing steps, we implemented a scaling factor \(\alpha\) for the standard deviations, as discussions with the domain experts have shown that this value is crucial for the validity of the simulation.
Above the process-specific reset bounds, the transition probabilities \(\mathcal{P}_{ss'}^a = Pr\left(s_{t+1} = s' \vert s_t = s, a_t = a\right)\) are defined in table~\ref{tab:manu_steps}. 
For example, applying MRF to a component with shape \(f_t\) results in a new shape \(f_{t+1} = \Delta f \cdot f_t\) with \(\Delta f \approx N(0.6, 0.01\alpha)\). 
Below the reset bounds, the process steps return samples from a belief state dependent normal distribution, please see table \ref{tab:manu_steps} for further information.

The reward \(r\) is calculated as a function of inverse processing time. The processing time, given in table \ref{tab:manu_steps} is a function of the processed area \(A\) and constants to express setup and tempering times.

The belief state \(B(s, h)= Pr(s_t = s \vert h_t=h)\) expresses our belief about the state \(s\). It depends on our initial belief about the state and the actions taken since. 
We use particles to approximate the belief state, each particle represents a sample from the belief state. This sample consists of values for the current belief about the current shape and roughness \((b_f, b_\sigma)\) and the number of cycles of the process steps that have been executed. As the number of cycles is deterministic, we only need particles for shape and roughness.
To generate the belief state for the history \(h\), we apply the actions from the history to the particles from the initial distribution.

\subsection{Calibrating the simulation}
\begin{figure*}[h]
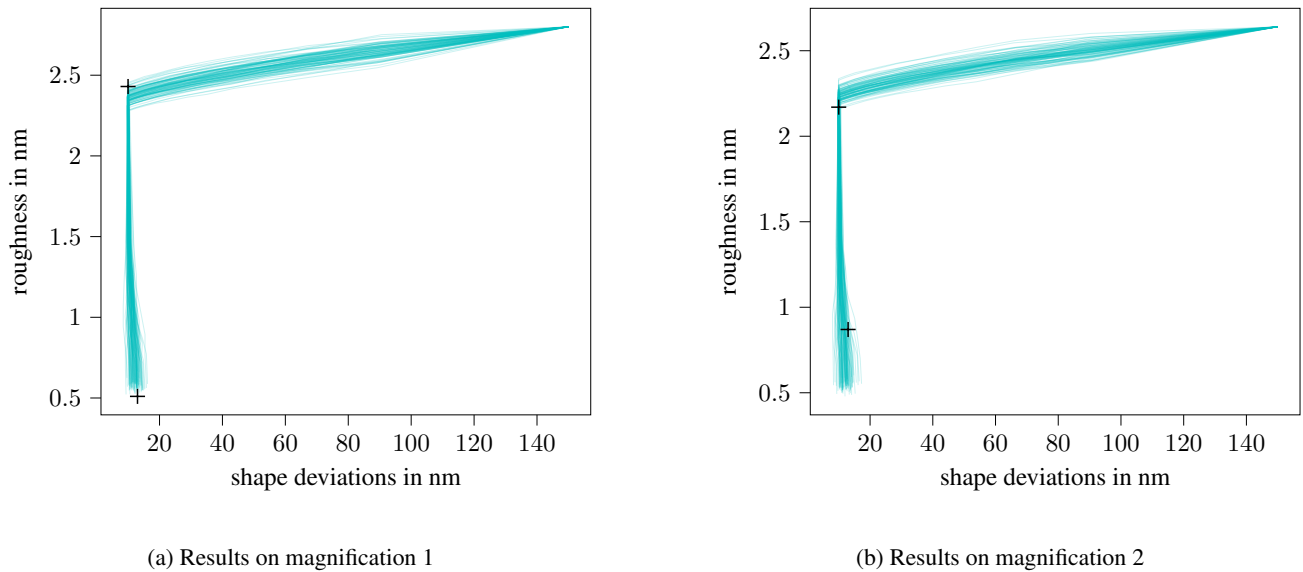

     \centering
     \begin{subfigure}[b]{0.45\textwidth}
         \centering
         \input{validation0.tex}
         \caption{Results on magnification 1}
     \end{subfigure}
     \hfill
     \begin{subfigure}[b]{0.45\textwidth}
         \centering
         \input{validation1.tex}
         \caption{Results on magnification 2}
     \end{subfigure}
        \caption{Trajectories in the shape-roughness-space for a single mirror with two roughness magnifications. Cyan: trajectories obtained by using the calibrated parameters, black crosses: the empirical data. The initial states correspond to the starting points of the trajectories in the upper right corners.}
        \label{fig:calibration}
\end{figure*}
We calibrate the simulation parameters of the MRF and CCP2 steps using results from real life. The sample was processed using a chain of MRF, CCP2 and CCP3 steps and measured after MRF and CCP3 using interferometry for the shape and white-light interferometry (WLI) for the roughness. WLI is comparable in precision and time with SLS. The roughness was measured using two different magnifications which we assume to be independent. In the real-life data, only the order of the processes (MRF - WLI - interferometry - CCP2 - CCP3 - WLI - interferometry), but not the number of manufacturing cycles was given.

In the calibration, we use the number of cycles and the mean values of the normal distributions as free parameters. For these, we solve a constrained optimization problem, minimizing the mean, (e.g. the mean over the two magnifications) relative deviations between the real-life processing and simulation outcomes. In consultation with the process engineers, we defined constraints (e.g., the improvement decreases with an increasing number of cycles) and bounds on the parameters.

Figure \ref{fig:calibration} shows the calibration result. 
With the calibrated parameters, we generate 100 trajectories through the roughness-shape-space, beginning from a fixed initial belief state in the upper right of the plot.
As black crosses, we give the measured calibration values. Except for the roughness parameter on magnification 2 (see figure \ref{fig:calibration}(b)), we achieve a good approximation to the measurement. Figure \ref{fig:calibration} also illustrates that MRF primarily improves the shape, while CCP steps change the surface roughness.
For the other process steps, the changes of the process parameters were defined by the process engineers.

\section{Optimization}
Given the particle-based simulator, we are able to optimize the manufacturing process. Our goal is to find a sequence of actions, such that all particles reach a user-defined shape and roughness in minimum average production time.
\subsection{MCTS for POMDP, POMCP}
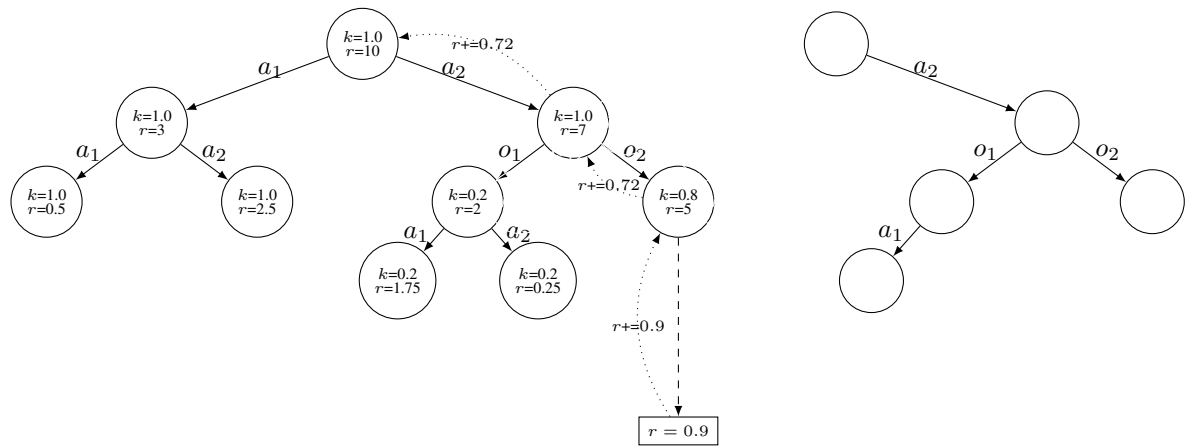
\begin{figure*}
\centering
    \input{tree1}
    \caption{
    A search tree with two actions and two observations and its transformation to an optimal branched tree of actions. 
    On the left, we see the search tree with the nodes corresponding to the belief states and the edges corresponding to actions $a$ or observations $o$. In the nodes, the share of all particles $k$ (relative to the amount of particle in the root node) and the rewards $r$ before a rollout is given. After a rollout in the rightmost node (dashed line), the reward $r$ is propagated back through the tree; the reward is weighted by the share $k$ of particles in the nodes (dotted lines).
    On the right, we see the transformed tree after evaluating all rewards.
}
    \label{fig:transform_tree}
 \end{figure*}
We follow the approach of \cite{Silver2010POMDP}, POMCP, an MCTS-algorithm to optimize POMDPs with large state spaces. It searches for an optimal sequence of actions, that moves the agent from a given intial belief state to a final belief state such that the agent's return is maximal. 

For this search POMCP iteratively forms and exploits a search tree. The edges in the search tree correspond to the actions or observations of the agent. 
In our application, only measurement steps return observations. Therefore, we omit the potential observation edges for the processing steps.

The search tree is constructed such that for each node there is only one path that leads to the root node. This backward path corresponds to one possible history \(h\) of the actor.
Hence the nodes of the search tree also represent the belief states that correspond to the history \(h\).
The belief states are approximated by a particle filter: in each node, \(K\) particles represent the actual belief state.
In our approach, the number of particles is not constant for all nodes. The particles of an observation node are distributed among its child nodes, such that the number of particles in a child  is proportional to the probability of the observation that corresponds to this child node, see figure~\ref{fig:transform_tree}.

To iteratively form the search tree, POMCP uses the algorithm Upper Confidence Bounds applied to Trees (UCT). 
UCT exploits the idea that in order to find the optimal action sequence we do not need to build the complete search tree: we are able to identify edges that potentially lead to leaves corresponding to runs with high reward. 
So instead of building the complete search tree at once, UCT iteratively adds new edges to an existing partial search tree. 
UCT uses two policies: the tree policy and the default policy. %Until a budget is met, 
The tree policy is used to select the most urgent node for examination, trading off exploration and exploitation by judging the number of visits and the gained reward in the node. If a node is selected by the tree policy, it is evaluated by the default policy: this means to draw a random or biased sequence of actions following the selected node, to calculate the return for this sequence and to propagate it back. As the number of particles in the nodes is not constant, we have to correct the return. An action leading to observations receives a return proportional to the rewards of its' children, linearly weighted by the share of particles in each child.

For application, we have to transform the partial search tree to an optimal chain of actions.
POMCP is an online algorithm: starting from the root node, it picks the action leading to the node with maximum mean return, executes it and immediately receives an observation. 
This executed action or observation moves the agent to a new node. Therefore, there is no need to continue building the tree for alternative nodes (nodes that we might have reached if we would have taken another action).
In our use case, we have no access to a real system, therefore we have to take all possible observations into account. As long as the observation tree is discrete, each possible observation defines a branch. Figure \ref{fig:transform_tree} shows this process. While actions are selected, the observations are exogenous, so all possible observations are stored.

\subsection{Applying POMCP in high precision manufacturing}
To apply POMCP in our production problem, we have to make some adaptions. On the one hand, our space of observations $O$ is continuous, so we have to find an appropriate discretization. On the other hand, we integrate expert-knowledge-based heuristics into the default policy, so it is easier to reach a terminal state.
\subsubsection{Binning measurement results}
For our application, the observation space is continuous and we have to discretize it.
At first, measurement results are drawn for all particles. Then, we apply a user-defined binning to these particles. This expresses the role of measurements in the real-world application, where further manufacturing steps depend on the measurement results. If a measurement step is reached in the default or tree policy, one of the corresponding observations is chosen randomly, based on the ratio of particles in the bin. 

\subsubsection{Using expert knowledge to define the default policy}
We use expert knowledge to define the default policy \cite{Browne2012survey, Silver2009MonteCarloSB} to reach the optimization goal faster than by picking random actions. Without these adaptions, the default policy is unlikely to arrive at a terminal state. This is due to three points: First, the terminal state can only be reached after measuring both quality metrics. In real world manufacturing, these final two obligatory steps are part of the final quality control. Second, some process steps (e.g. CCP2, CCP3, MRF) reset the quality metrics if applied below their reset bounds. This induces disadvantageous processing loops, as the state is reset to a state that has already been reached. This is very crucial especially for MRF, that resets the roughness to a rather high value of \SI{2}{nm}. Third, CCP3 is a final processing step. It is only useful if the roughness and the shape are are sufficiently small. In addition, only one cycle of CCP3 can be executed.

We face these points by introducing heuristics into the default policy. Regarding the obligatory final measurement steps, our default policy initiates these two measurements as soon as all particles of the belief state correspond to quality metrics within the target interval. 

To deal with the reset bounds and the special requirement for CCP3, we use the following heuristics: if the particle values are above the reset bound for MRF, we use either MRF or CCP2 with defined probabilities, depending on the belief state values of the particles.
Below the reset bound, we prefer a chain of actions consisting of multiple cycles of CCP2, followed by one cycle of CCP3. Please refer to the appendix for further details on the heuristics. 
\subsection{Results of POMCP}
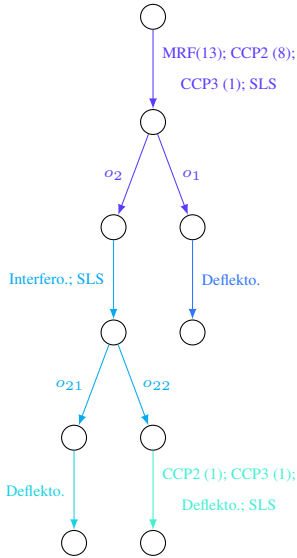
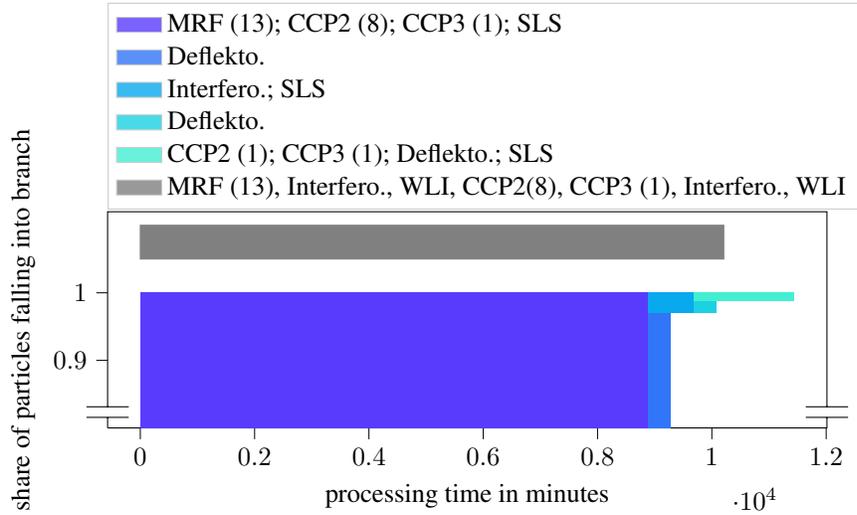
\begin{figure*}
     \centering
     \begin{subfigure}[b]{0.25\textwidth}
         \centering
            \input{final_abstract_tree}
         \caption{The optimal manufacturing plan represented by a branched of actions.
         }%Final branched tree, skipping states between processing actions.}
     \end{subfigure}
     \hfill
     \begin{subfigure}[b]{0.65\textwidth}
         \centering
            \input{final_tree}
         \caption{The share of particles of the optimized solution (blues) and the real-life benchmark (gray) over the processing time.}
     \end{subfigure}
    \centering
    \caption{Given an initial population of mirrors, the noise in the manufacturing process leads to different process chains.
    The optimal manufacturing plan (a) is represented as a branched tree of actions: The actions and observation outcomes are represented by the edges; consecutive processing steps are represented by one edge. The tree branches according to the possible observation outcomes. The optimal plan consists of three alternative paths for manufacturing the mirror, the alternatives are chosen according to the results of the measuring actions.
 The distribution of the processing time (b) is determined by the probability of choosing one of the three branches, given by the share of particles falling into each branch.
    }
    \label{fig:final_process}
\end{figure*}
To evaluate our approach, we compare process chains optimized by POMCP with data from the real-life manufacturing of a flat mirror. 
We are given initial values of shape accuracy \(f_0\) = \SI{150}{nm} and a roughness \(\sigma_0\) = \SI{2.8}{nm} and target values of \(f_{T}\) = \SI{13}{nm} and roughness \(\sigma_{T}\) = \SI{0.5}{nm}.

In real-life, the mirror was processed by a chain of MRF, CCP2 and CCP3 steps. After all MRF cycles and after the CCP3 processing, roughness and shape were evaluated by interferometry and WLI. Due to the high processing costs, we are given only one real-life process chain, so we only see one realization of the potential distribution of processing outcomes.

Figure \ref{fig:final_process} shows a comparison of the optimized result with results from real-life processing. The processing time is given on the $x$-axis, while the $y$-axis shows the share of particles that fall into a specific branch of the tree of actions. The largest share of all particles falls into a process branch that corresponds to the real-life processing chain, but skips the costly measurements after the MRF steps. The rest of the particles is further processed depending on actual measurement results, e.g., the bins, the particles fall into. After the main process chain ending with a SLS measurement step, a share of the particles with large roughness is measured and processed by additional CCP2 and CCP3 steps. Another share is only remeasured, until all particles fall into the target region. This behaviour takes the measurement noise into account.

Overall, the improvement of using MCTS is not only the optimized process chain, but also recommended actions for different measurement results. This helps the process engineers not only during production but also when planning the manufacturing of new mirrors.

%--------------------------------------------------------
\section{Conclusion and Outlook}
In this paper, we have described a comprehensive approach for the use of POMCP in high precision manufacturing. The approach makes use of different steps, building on each other. The first step is the implementation of a simulator based on expert knowledge and its calibration with real world data. 

As a second step, we fit the manufacturing process into the POMDP framework. We distinguish between two types of manufacturing steps: processing steps change the belief state, but do not return observations, whilst measurements report the belief state by observations. 

To optimize the production process in a third step for a minimum processing time, we build on the work of \cite{Silver2010POMDP}, for MCTS in large POMDPs. Our optimization goal is a minimum processing time for a given product tolerance. We make adaptions on the original algorithm to work offline (e.g. no pruning when receiving real world observations), to deal with continuous measurement results (e.g. binning of the measurements) and to encourage convergence (e.g. adapting the default policy).

In a fourth step, we compare our optimized results with results from a real-life production process. The optimized result skips costly intermediate measurements and returns recommended actions dependent on the actual measurement results.

Overall, the approach provides insights into the optimization of complex stochastic manufacturing processes such as high precision manufacturing processes. Based on the already good results, we see several opportunities for expansion: to use a finer simulation that returns local metric values instead of a global value for the whole mirror; to take into account additional quality metrics and finally to optimize for the machining capacities instead of the machining time.

\section*{Acknowledgements}
This work was jointly developed by the Fraunhofer Lighthouse
project SWAP, the Fraunhofer Center for Machine Learning within
the Fraunhofer Cluster for Cognitive Internet Technologies and has been partly funded by the Federal Ministry of 
Education and Research of Germany as part of the competence center for 
machine learning ML2R (01IS18038B).

\bibliography{rl4rl_ws}
\bibliographystyle{icml2021}

\appendix
\section{Changes to the default policy}
We insert the following heuristics: if the roughness (observed or belief state) is above \SI{2}{nm}, the default policy chooses either a CCP2 or a MRF step with defined probabilities, depending on the ratio of objective fulfillment: \(p_{\text{MRF}} = \frac{b_\sigma}{\targetroughness} / \frac{b_{f}}{\targetform}, p_{\text{CCP2}} = 1 - p_{\text{MRF}}\). 
If the roughness is below \SI{2}{nm}, the use of MRF is forbidden. To decide, which process to use, we take advantage of the expert knowledge about the expected changes by the processing steps; it is most advantageous to perform multiple cycles of CCP2 followed by one CCP3 step. In expectation, each cycle of CCP2 improves the roughness by 10~\% and increases the shape deviation by the same amount. Respectively, we have for CCP3 the values 25~\% and 2.5~\%. We search for a minimal number of CCP2 cycles \(n\), such that the following equations hold:
\begin{equation}
    \begin{split}
         b_\sigma \cdot 0.75 \cdot 0.9^n \leq &\ \targetroughness \\
    b_f \cdot 1.025 \cdot 1.1^n \leq &\ \targetform     
    \end{split}
\end{equation}
Combining this we theoretically can start a chain of \(n\) CCP2 cycles, whenever:
\begin{equation}
   \log_{0.9}\left(\frac{\targetroughness}{b_\sigma\cdot 0.75}\right)< \log_{1.1}\left(\frac{\targetform}{b_f\cdot 1.025}\right)
   \label{eq:start_chain}
\end{equation}
But, as the processing results are randomized, with standard deviation scaled with the randomness factor \(\alpha\), this heuristic is too conservative. Therefore, we multiply the left hand side of equation (\ref{eq:start_chain}) by \((1+\alpha)\).
If equation (\ref{eq:start_chain}) is not true, but the shape is below the threshold value of \SI{2}{nm}, the default policy makes use of either CCP1 or MRF. Both methods are disadvantageous, as CCP1 has a bad ratio of improvement to processing time, while MRF resets the form.

\end{document}

%% file: tree1.tex
\usetikzlibrary{trees,arrows}
%\begin{document}
\begin{tikzpicture}[scale=0.7,edge from parent/.style={draw,-latex},level/.style={sibling distance = 8cm/#1}]
  \node at (1,0) [circle,draw,align=center,execute at begin node=\setlength{\baselineskip}{-5ex}](q0) {\tiny $k$=1.0\\\tiny$r$=10} 
    child {node [circle,draw,align=center,execute at begin node=\setlength{\baselineskip}{-5ex}] (q1) {\tiny $k$=1.0\\\tiny$r$=3}  
        child {node [circle,draw,align=center,execute at begin node=\setlength{\baselineskip}{-5ex}] (q11) {\tiny $k$=1.0\\\tiny$r$=0.5}
            %child [white]{node [yshift=-2cm,rectangle,draw] (r11) {\tiny$r=0.01$}
            %edge from parent [dashed] {}
            %}        
        edge from parent node[left,near start] {$a_1$}}
        child {node [circle,draw,align=center,execute at begin node=\setlength{\baselineskip}{-5ex}] (q12) {\tiny $k$=1.0\\\tiny$r$=2.5}
            %child [white]{node [yshift=-2cm,rectangle,draw] (r12) {\tiny$r=0.05$}
            %edge from parent [dashed] {}
            %}
        edge from parent node[right,near start] {$a_2$}}
    edge from parent node[left,near start] {$a_1$}}
    child {node [circle,draw,align=center,execute at begin node=\setlength{\baselineskip}{-5ex}] (q2) {\tiny $k$=1.0\\\tiny$r$=7}
        child {node [circle,draw,align=center,execute at begin node=\setlength{\baselineskip}{-5ex}] (q21) {\tiny $k$=0.2\\\tiny$r$=2}
            child {node [circle,draw,align=center,execute at begin node=\setlength{\baselineskip}{-5ex}] (q212) {\tiny $k$=0.2\\\tiny$r$=1.75}
                %child [white]{node [yshift=-0.95cm,rectangle,draw] (r212) {\tiny$r=0.8$}
                %edge from parent [dashed] {}
                %}
            edge from parent node[left,near start] {$a_1$}}
            child {node [circle,draw,align=center,execute at begin node=\setlength{\baselineskip}{-5ex}] (q222) {\tiny $k$=0.2\\\tiny$r$=0.25}
                %child [white]{node [yshift=-0.95cm,rectangle,draw] (r222) {\tiny$r=0.2$}
                %edge from parent [dashed] {}
                %}
            edge from parent node[right,near start] {$a_2$}}
        edge from parent node[left,near start] {$o_1$}}
        child {node [circle,draw,align=center,execute at begin node=\setlength{\baselineskip}{-5ex}] (q22) {\tiny $k$=0.8\\\tiny$r$=5}
            child {node [yshift=-2cm,rectangle,draw] (r22) {\tiny$r=0.9$}
            edge from parent [dashed] {}}
        edge from parent node[right,near start] {$o_2$}}
    edge from parent node[right,near start] {$a_2$}};
    
%\draw[dotted, bend left, ->,>=latex](r11) to node[] {\tiny$r$+=$0.01$} (q11);
%\draw[dotted, bend right, ->,>=latex](q11) to node[] {} (q1);
%\draw[dotted, bend left, ->,>=latex](q1) to node[] {\tiny$r$+=$0.06$} (q0);
%\draw[dotted, bend left, ->,>=latex](r12) to node[] {\tiny$r$+=$0.05$} (q12);
%\draw[dotted, bend left, ->,>=latex](q12) to node[] {} (q1);
%\draw[dotted, bend left, ->,>=latex](r222) to node[] {\tiny$r$+=$0.2$} (q222);
%\draw[dotted, bend left, ->,>=latex](r212) to node[] {\tiny$r$+=$0.8$} (q212);
%\draw[dotted, bend left, ->,>=latex](q222) to node[] {} (q21);
%\draw[dotted, bend right, ->,>=latex](q212) to node[] {} (q21);
\draw[dotted, bend right, ->,>=latex](q2) to node[] {\tiny$r$+=$0.72$} (q0);
\draw[dotted, bend left, ->,>=latex](q22) to node[] {\tiny$r$+=$0.72$} (q2);
\draw[dotted, bend left, ->,>=latex](r22) to node[] {\tiny$r$+=$0.9$} (q22);
%\draw[dotted, bend right, ->,>=latex](q21) to node[] {\tiny$r$+=$0.2$} (q2);

  \node at (10,0) [text=white,circle,draw](s0) {\tiny $k$=1.0} 
    child [white] {node [circle,draw] (s1) {\tiny $k$=1.0}  
        child {node [text=white,circle,draw] (s11) {\tiny $k$=1.0}
%            child {node [yshift=-2cm,rectangle,draw] (r12) {\tiny$r=0.01$}
%            edge from parent [dashed] {}}        
        edge from parent node[left,near start] {$a_1$}}
        child {node [text=white,circle,draw] (s12) {\tiny $k$=1.0}
%            child {node [yshift=-2cm,rectangle,draw] (r12) {\tiny$r=0.05$}
%            edge from parent [dashed] {}}
        edge from parent node[right,near start] {$a_2$}}
    edge from parent node[left,near start] {$a_1$}}
    child {node [text=white,circle,draw] (s2) {\tiny $k$=1.0}
        child {node [text=white,circle,draw] (s21) {\tiny $k$=1.0}
            child {node [text=white,circle,draw] (s212) {\tiny $k$=1.0}
%                child {node [yshift=-0.5cm,rectangle,draw] (r12) {\tiny$r=0.8$}
%                edge from parent [dashed] {}}
            edge from parent node[left,near start] {$a_1$}}
            child [white]{node [circle,draw] (s222) {\tiny $k$=1.0}
%                child {node [yshift=-0.5cm,rectangle,draw] (r12) {\tiny$r=0.9$}
%                edge from parent [dashed] {}}
            edge from parent node[right,near start] {$a_2$}}
        edge from parent node[left,near start] {$o_1$}}
        child {node [text=white,circle,draw] (s22) {\tiny $k$=1.0}
%            child {node [yshift=-2cm,rectangle,draw] (r12) {\tiny$r=0.9$}
%            edge from parent [dashed] {}}
        edge from parent node[right,near start] {$o_2$}}
    edge from parent node[right,near start] {$a_2$}};
\end{tikzpicture}
%\end{document}

%% file: final_abstract_tree.tex
\usetikzlibrary{trees,arrows}
\begin{tikzpicture}[scale=0.7,edge from parent/.style={draw,-latex},
level distance=2cm
]
\definecolor{color0}{rgb}{0.350980392156863,0.231947641453898,0.993158666136636}
\definecolor{color1}{rgb}{0.194117647058824,0.462203883540313,0.971281031916114}
\definecolor{color2}{rgb}{0.0372549019607843,0.664540178707858,0.934679767321111}
\definecolor{color3}{rgb}{0.111764705882353,0.819740482907221,0.886773685920062}
\definecolor{color4}{rgb}{0.268627450980392,0.934679767321111,0.823252948157587}
  
  \node [circle,draw](q0) {} 
    child [color0]{node [black][circle,draw] (q2) {}
        child [color0]{node [black][circle,draw] (o2){}
            child [color2]{node [black] [circle,draw] (q21) {}
                child [color2]{node [black] [circle,draw] (o21) {}
                    child [color3]{node [black] [circle,draw] (q212) {}
%                        child [white]{node [circle,draw] (q2121) {}
%                        edge from parent node[left] {}}
                    edge from parent node[left] {\tiny Deflekto.}}
                edge from parent node[left] {\tiny$o_{21}$}}
                child [color2]{node [black] [circle,draw] (o22) {}
                    child [color4]{node [black][circle,draw] (q222) {}
                    edge from parent node[align=center,right,] {\tiny CCP2 (1); CCP3 (1);\\\tiny Deflekto.; SLS}}
                edge from parent node[right] {\tiny$o_{22}$}}
            edge from parent node[left] {\tiny Interfero.; SLS}}
        edge from parent node[align=center,left,] {\tiny$o_2$}}
        child[color0]{node [black][circle,draw] (o1){}
            child [color1]{node [black][circle,draw] (q22) {}
            edge from parent node[right] {\tiny Deflekto.}}
        edge from parent node[align=center,right,] {\tiny$o_1$}}
    edge from parent node[align=center,right,] {\tiny MRF(13); CCP2 (8);\\\tiny CCP3 (1); SLS}};

\end{tikzpicture}

%% file: final_tree.tex
% This file was created by tikzplotlib v0.9.8.
\usetikzlibrary{trees,arrows}
\begin{tikzpicture}

\definecolor{color0}{rgb}{0.350980392156863,0.231947641453898,0.993158666136636}
\definecolor{color1}{rgb}{0.194117647058824,0.462203883540313,0.971281031916114}
\definecolor{color2}{rgb}{0.0372549019607843,0.664540178707858,0.934679767321111}
\definecolor{color3}{rgb}{0.111764705882353,0.819740482907221,0.886773685920062}
\definecolor{color4}{rgb}{0.268627450980392,0.934679767321111,0.823252948157587}

\begin{axis}[
width=1.\textwidth,
height=0.4\textwidth,
legend cell align={left},
legend style={
  fill opacity=0.8,
  draw opacity=1,
  text opacity=1,
  at={(0,1.01)},
  anchor=south west,
  draw=white!80!black
},
tick align=outside,
tick pos=left,
x grid style={white!69.0196078431373!black},
xlabel={processing time in minutes},
xmin=-572.066264951337, xmax=12013.3915639781,
xtick style={color=black},
y grid style={white!69.0196078431373!black},
ylabel={share of particles falling into branch},
axis y discontinuity=parallel,
ymin=0.8, ymax=1.12,
ytick = {0.9,1.},
ytick style={color=black}
]
\path [fill=color0]
(axis cs:0,1)
--(axis cs:0,0)
--(axis cs:8890.07672210187,0)
--(axis cs:8890.07672210187,1)
--(axis cs:8890.07672210187,1)
--(axis cs:0,1)
--cycle;
\addlegendimage{area legend, fill=color0}
\addlegendentry{   MRF (13); CCP2 (8); CCP3 (1);  SLS}

\path [fill=color1]
(axis cs:8890.07672210187,0.97)
--(axis cs:8890.07672210187,0)
--(axis cs:9282.00518815633,0)
--(axis cs:9282.00518815633,0.97)
--(axis cs:9282.00518815633,0.97)
--(axis cs:8890.07672210187,0.97)
--cycle;
\addlegendimage{area legend, fill=color1}
\addlegendentry{  Deflekto.}

\path [fill=color2]
(axis cs:8890.07672210187,1)
--(axis cs:8890.07672210187,0.97)
--(axis cs:9689.3201108704,0.97)
--(axis cs:9689.3201108704,1)
--(axis cs:9689.3201108704,1)
--(axis cs:8890.07672210187,1)
--cycle;
\addlegendimage{area legend, fill=color2}
\addlegendentry{Interfero.;  SLS}

\path [fill=color3]
(axis cs:9689.3201108704,0.988)
--(axis cs:9689.3201108704,0.97)
--(axis cs:10081.2485769249,0.97)
--(axis cs:10081.2485769249,0.988)
--(axis cs:10081.2485769249,0.988)
--(axis cs:9689.3201108704,0.988)
--cycle;
\addlegendimage{area legend, fill=color3}
\addlegendentry{Deflekto.}

\path [fill=color4]
(axis cs:9689.3201108704,1)
--(axis cs:9689.3201108704,0.988)
--(axis cs:11441.3252990267,0.988)
--(axis cs:11441.3252990267,1)
--(axis cs:11441.3252990267,1)
--(axis cs:9689.3201108704,1)
--cycle;
\addlegendimage{area legend, fill=color4}
\addlegendentry{  CCP2 (1); CCP3 (1); Deflekto.;  SLS}

\path [draw=white!50.1960784313725!black, fill=white!50.1960784313725!black]
(axis cs:0,1.1)
--(axis cs:0,1.05)
--(axis cs:10208.4867775371,1.05)
--(axis cs:10208.4867775371,1.1)
--(axis cs:10208.4867775371,1.1)
--(axis cs:0,1.1)
--cycle;
\addlegendimage{area legend, fill=white!50.1960784313725!black}
\addlegendentry{MRF (13), Interfero., WLI, CCP2(8), CCP3 (1), Interfero., WLI}

\end{axis}
\end{tikzpicture}